\pdfoutput=1

\documentclass[11pt]{article}

\usepackage{acl}

\usepackage{times}
\usepackage{latexsym}
\usepackage{newfloat}
\usepackage{listings}
\usepackage{algorithm}
\usepackage{algorithmic}
\usepackage{dsfont}
\usepackage{threeparttable}
\usepackage{booktabs}
\usepackage{multicol}
\usepackage{multirow}
\usepackage{diagbox}
\usepackage{amsmath}
\usepackage{float}
\usepackage{caption}
\usepackage{subcaption}
\usepackage{graphicx}
\usepackage[T1]{fontenc}
\usepackage[utf8]{inputenc}
\usepackage{microtype}

\title{Two-Level Supervised Contrastive Learning for Response Selection in Multi-Turn Dialogue}

\author{
Wentao Zhang \\
University of Illinois at Urbana-Champaign \\
\texttt{wentao4@illinois.edu} \\
\And
Shuang Xu \\
ByteDance AI Lab\\
\texttt{xushuang.0420@bytedance.com}\\
\AND
Haoran Huang \\
ByteDance AI Lab\\
\texttt{huanghaoran@bytedance.com} \\
}

\begin{document}
\maketitle
\begin{abstract}
Selecting an appropriate response from many candidates given the utterances in a multi-turn dialogue is the key problem for a retrieval-based dialogue system.  Existing work formalizes the task as matching between the utterances and a candidate and uses the cross-entropy loss in learning of the model. This paper applies contrastive learning to the problem by using the supervised contrastive loss.  In this way, the learned representations of positive examples and representations of negative examples can be more distantly separated in the embedding space, and the performance of matching can be enhanced. We further develop a new method for supervised contrastive learning, referred to as two-level supervised contrastive learning, and employ the method in response selection in multi-turn dialogue. Our method exploits two techniques:  sentence token shuffling (STS) and sentence re-ordering (SR) for supervised contrastive learning. Experimental results on three benchmark datasets demonstrate that the proposed method significantly outperforms the contrastive learning baseline and the state-of-the-art methods for the task.
\end{abstract}

\begin{figure*}[htbp]
\centering
\includegraphics[width=0.8\textwidth, trim={0cm 0cm 0cm 0.1cm}, clip]{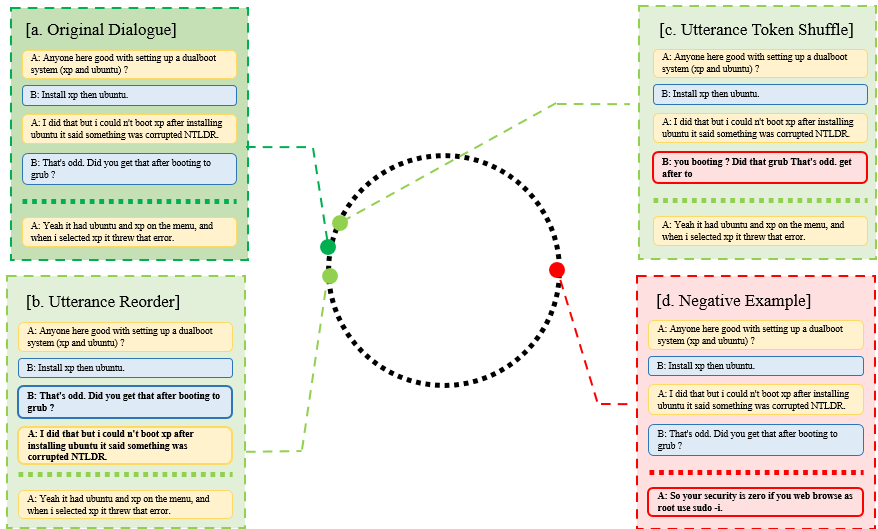} 
\caption{Example of two augmentation approaches that we integrated into contrastive learning model. The distances of hidden representations between the original dialogue and augmented example under the cosine similarity metric are expected to be closer than those between the original ones and corresponding negative examples.}
\label{main}
\end{figure*}

\section{Introduction}
Recent years have observed a significant surge in research on intelligent dialogue systems that can communicate with humans in natural language. In this paper, we focus on response selection in a multi-turn dialogue, in which the system selects the most appropriate response from a set of candidates given the utterances in the dialogue history.


Nowadays, pre-trained language models such as BERT \cite{devlin-etal-2019-bert}, RoBERTa \cite{liu2020roberta} and ELECTRA \cite{Clark2020ELECTRA:} have become fundamental technologies of NLP. Naturally, one can formalize response selection as a matching problem and employ a pre-trained language model to conduct the task, where the input consists of a candidate of response and the utterances in the dialogue history.  There is still room for improvement, however. Usually, the cross-entropy loss is used in training of the model, which is sensitive to noisy labels~\cite{noisy} and incapable of creating a large margin between positive and negative examples~\cite{margin1, margin-2}. In other words, the model learned with the cross entropy may not be strong enough to capture the features that can distinctively separate the positive examples and negative examples. \citet{1640964} originally propose exploiting contrastive learning as a way of self-supervised learning, in which positive and negative examples are automatically created and the contrastive loss is used.
Empirical results have proved the effectiveness of contrastive learning in computer vision~\cite{wu2018unsupervised, bachman2019amdim, tian2019contrastive,he2019moco, chen2020simple, chen2020big, chen2020mocov2} and NLP~\cite{gao2021simcse}. Contrastive learning thus can be naturally employed to address the problem described above. Particularly, one can use the method of SimCSE to perform the task of response selection. We elaborate these related work in Appendix.



We propose a new method of supervised contrastive learning, referred to as two-level supervised contrastive learning. We linearly combine the contrastive loss and the cross-entropy loss as the loss function. In addition, we use two techniques to augment the input text data at both the token level and sentence level for positive examples, namely sentence token shuffling (STS) and sentence re-ordering (SR), to create positive examples. For negative examples, we actively leverage the high quality mis-matched candidate provided by dataset. Our method uses BERT as the pre-trained language model, takes the utterances and a candidate as input, and fine-tunes the model using the loss function mentioned above. In this way, we can create a model that can effectively separate positive examples from negative examples in the embedding space.

We test our method on three benchmark datasets for retrieval-based dialogue: Ubuntu~\cite{lowe-etal-2015-ubuntu}, Douban~\cite{wu-etal-2017-sequential}, and E-Commerce Dataset~\cite{zhang-etal-2018-modeling}. 
Ablation study also shows the necessity of the components of our method, i.e., the loss function and the two techniques. 

Our contributions are as follows.
\begin{itemize}    
    \item We propose a new supervised contrastive learning method using the contrastive loss supported by two augmentation techniques.
    \item We propose employing the method for enhancing the performance of response selection in multi-turn dialogue.
\end{itemize}

Experimental results show that our method outperforms existing state-of-the-art methods by a large margin.

\section{Model and Approach}

Suppose that there is a multi-turn dialogue dataset $\mathcal{D}=(c_i, r_i, y_i)^N_{i=1}$, where $c_i = \{u_{i, 1}, u_{i, 2}, u_{i, 3}, \cdots, u_{i, l_i}\}$ is the context which consists of $l_i$ utterances, $r_i$ is the response for $c_i$ and $y_i \in \{0, 1\}$ is the ground truth label. We need to learn a model $F$ from $\mathcal{D}$ so that for any given context $c$ and response candidate $r$, $F(c, r) \in [0, 1]$ can measure the matching degree between $c$ and $r$.

Given a pre-trained language model $M$, we first perform domain adaptive pre-training using $\mathcal{D}$ to make the representation more domain-relevant and applicable to downstream tasks (see Appendix A.1), then we fine-tune $M$ on $\mathcal{D}$ using a contrastive fine-tuning framework.

\subsection{Contrastive Fine-tuning Framework} 
We apply contrastive learning in the fine-tuning process to leverage label information. There are three major components in our framework: a data augmentation module that generates different views for input samples, an encoder $M$ that computes representations for each input text and a loss layer on top of the encoder that combines binary cross-entropy loss and contrastive loss.
 

\subsection{Data Augmentation Strategies}


\subsubsection{Sentence Token Shuffling (STS)} Instead of shuffling all the tokens in the input sequences like \cite{yan2021consert}, we randomly select one utterance in the context and shuffle all tokens to form a new training example. Part c of Figure \ref{main} illustrates how to perform sentence token shuffling.


\subsubsection{Sentence Reordering (SR)}  In previous work \cite{Zhang2021StructuralPF}, researchers let the model detect the sentence in the wrong place or predict the correct order of the input sequence. In this work, we randomly choose two utterances in the context and exchange their positions to get the augmentation. This approach enhances PrLMs' sentence level modeling ability. To make the augmented example more reasonable, we never change the position of the last utterance in the dialogue context. Part b of Figure \ref{main} illustrates how to perform utterance reordering augmentation.

\subsubsection{Two-level Augmentation Approach (TL)} 
We propose a technique that combines token-level and sentence-level approaches to create noisy positive samples. Specifically, we duplicate the original batch twice to get STS and SR augmentations. By incorporating two augmentation approaches into contrastive learning, the model further improves the expressiveness of the representations.

\subsection{Supervised Contrastive Loss}
For each batch of samples $X$ with size $N$, we generate augmented batch $X^+$ as we described above, and retrieve corresponding negative batch $X^-$, which contains the same context but with misleading response provided by the dataset, to get $(X, X^+, X^-)$. Then we encode these batches with PrLMs to get hidden representations $(H, H^+, H^-)$. The supervised contrastive loss can be defined as
\begin{equation}
    \label{scl_loss}
    \mathcal{L}_{SCL} = -\sum_{i=1}^{N}\log\frac{e^{f(h_i, h_i^+) / \tau}}{Z(H, H^+, H^-)} 
\end{equation}
and
\begin{equation}
\begin{array}{l}
     Z(H, H^+, H^-) \\
     =  \sum_{j=1}^{N}\left(e^{f(h_i, h_j^+)/\tau} + e^{(f(h_i, h_j^-) + \alpha \mathds{1}_i^j)/\tau}\right)
\end{array}
\end{equation}where $h_i, h_i^+, h_i^-$ is the $i$th hidden representation of original, augmented and negative data batch, and $f$ is the similarity function as which we adopt cosine similarity in this work. $\mathds{1}_i^j \in \{0, 1\}$ is the indicator function that equals 1 if and only if $i=j$, and $\alpha$ is the penalty factor for hard negative samples.

Since the prevailing datasets have already provided true negative pairs, it may be unreasonable to treat these high-quality negative examples equally with other in-batch negatives. To fully leverage these hard negative, we add a penalty $\alpha$ to the similarity score of $h_i$ and $h_i^-$ so that the model can significantly contrast them.


Finally, we add the supervised contrastive loss to binary cross-entropy loss with  hyper-parameter $\lambda$ and get the final objective function:
\begin{equation}
    \mathcal{L} = \mathcal{L}_{SCL} + \lambda \mathcal{L}_{CE}
    \label{final_loss}
\end{equation}

\section{Experiments}
\subsection{Datasets}
We evaluate the proposed method on three public multi-turn response selection datasets, Ubuntu Dialogue Corpus V1 \citet{Xu2016IncorporatingLK}, Douban Corpus \cite{wu-etal-2017-sequential} and E-commerce Dialogue Corpus \cite{zhang-etal-2018-modeling}. Some statistics of these datasets are provided in Table \ref{data_profile} (see Appendix A.2). 

\subsection{Evaluation Metrics}

Following the previous work, we use $\mbox{Recall}_n@k$ or $R_n@k, k = \{1, 2, 5\}$ as evaluation metrics, where $n$ is the number of response candidates of each context and $k$ is the top-$k$ scored candidates among $n$ candidates. Mean average precision (MAP), mean reciprocal rank (MRR), and precision at one (P@1) are also used for Douban Corpus to evaluate the model's ability to retrieve multiple responses.

\begin{table*}[t]
\centering
 \resizebox{\textwidth}{!}{
\begin{threeparttable}

\begin{tabular}{l|l|ccc|cccccc|cccc} 
\toprule
\multirow{2}{*}{}  & \multirow{2}{*}{\diagbox{\textbf{Model Name}}{\textbf{Data}}} & \multicolumn{3}{c|}{E-Commerce} & \multicolumn{6}{c|}{Douban}  & \multicolumn{4}{c}{Ubuntu V1}  \\ 
\cline{3-15}
    & & $\text{\textbf{Recall}}_{10}@1$    & $\text{\textbf{Recall}}_{10}@2$ & $\text{\textbf{Recall}}_{10}@5$   & \textbf{MAP}   & \textbf{MRR}   & $\text{\textbf{P}}@1$  & $\text{\textbf{Recall}}_{10}@1$  & $\text{\textbf{Recall}}_{10}@2$ & $\text{\textbf{Recall}}_{10}@5$    & $\text{\textbf{Recall}}_2@1$ & $\text{\textbf{Recall}}_{10}@1$  & $\text{\textbf{Recall}}_{10}@2$  & $\text{\textbf{Recall}}_{10}@5$\\ 
\hline
\multirow{8}{*}{\begin{tabular}[c]{@{}l@{}}non\\PrLM-based\end{tabular}} & LSTM \cite{lowe-etal-2015-ubuntu} & 0.365 & 0.536 & 0.828 &  0.485 &  0.537 &  0.320 & 0.187 & 0.343 & 0.720 & - & 0.638 & 0.784 & 0.949  \\

    & SMN \cite{wu-etal-2017-sequential}  & 0.453  & 0.654 & 0.886  & 0.529 & 0.569 & 0.397  & 0.233  & 0.396   & 0.724 & 0.926  & 0.726  & 0.847  & 0.961    \\
    & DUA \cite{zhang-etal-2018-modeling}  & 0.501  & 0.700 & 0.921  & 0.551 & 0.599 & 0.421  & 0.243  & 0.421   & 0.780 & - & 0.752  & 0.868  & 0.962   \\
    & DAM \cite{zhou-etal-2018-multi}  & 0.526  & 0.727 & 0.933 & 0.550 & 0.601 & 0.427  & 0.254  & 0.410   & 0.757  & 0.938  & 0.767  & 0.874  & 0.969   \\
    & MRFN \cite{mrfn} & - & -   & -   & 0.571 & 0.617 & 0.448  & 0.276  & 0.435   & 0.783 & 0.945  & 0.786  & 0.886  & 0.976  \\
    & IMN \cite{imn}   & 0.621  & 0.797 & 0.964 & 0.570 & 0.615 & 0.433  & 0.262  & 0.452   & 0.789 & - & 0.794  & 0.889  & 0.974    \\
    & IoI \cite{tao-etal-2019-one}   & - & -   & -  & 0.573 & 0.621 & 0.444  & 0.269  & 0.451   & 0.786  & 0.947  & 0.796  & 0.894  & 0.974   \\
    & MSN \cite{yuan-etal-2019-multi}   & 0.606  & 0.770 & 0.937  & 0.587 & 0.632 & 0.470  & 0.295  & 0.452   & 0.788   & - & 0.800  & 0.899  & 0.978 \\ 
\hline
\multirow{9}{*}{\begin{tabular}[c]{@{}l@{}}PrLMs-based\end{tabular}} & BERT\cite{sa-bert}  & 0.610  & 0.814 & 0.973& 0.591 & 0.633 & 0.454  & 0.280  & 0.470   & 0.828   & 0.950  & 0.808  & 0.897  & 0.975   \\
& BERT-VFT \cite{bert-vft}   & 0.717 &  0.884 &  0.986  & - & - & - & - & - & - & 0.969 & 0.867 & 0.939 & 0.987\\
    & SA-BERT \cite{devlin-etal-2019-bert} & 0.704  & 0.879 & 0.985  & 0.619 & 0.659 & 0.496  & 0.313  & 0.481   & 0.847 & 0.965  & 0.855  & 0.928  & 0.983   \\
    & UMS-BERT+\cite{Whang_Lee_Oh_Lee_Han_Lee_Lee_2021} & 0.762  & 0.905 & 0.986  & 0.625  & 0.664  & 0.499 & 0.318  & 0.482   & 0.858  & - & 0.875  & 0.942  & 0.988 \\
    & ELECTRA\cite{Whang_Lee_Oh_Lee_Han_Lee_Lee_2021}  & 0.607  & 0.813 & 0.960 & 0.599 & 0.643 & 0.471  & 0.287  & 0.474   & 0.831  & 0.960  & 0.845  & 0.919  & 0.979  \\
    & UMS-ELECTRA+\cite{Whang_Lee_Oh_Lee_Han_Lee_Lee_2021} & 0.707  & 0.853 & 0.974  & 0.623 & 0.663 & 0.492  & 0.307  & 0.501*  & 0.851  & - & 0.875  & 0.941  & 0.988  \\
    & MDFN\cite{Liu2021FillingTG}  & 0.639  & 0.829 & 0.971  & 0.624 & 0.663 & 0.498  & 0.325  & 0.511   & 0.855  & 0.967  & 0.866  & 0.932  & 0.984  \\
    & BERT-SL\cite{Xu2021LearningAE} & 0.776 & 0.919  & 0.991  & -   & -   & - & - & - & -  & 0.975 & 0.884 & 0.946 & 0.990   \\ 
    & BERT-FP\cite{han-etal-2021-fine} & 0.870 & 0.956  & 0.993 & 0.644  & 0.680  & 0.512 &0.324- & 0.542 & 0.870    & - & \textbf{0.911} & \textbf{0.962} & \textbf{0.994}  \\ 
\hline
\multirow{2}{*}{\begin{tabular}[c]{@{}l@{}}Our Model\end{tabular}}
    & $\mbox{BERT}^+_{TL}$  &   \textbf{0.927} & \textbf{0.974} & \textbf{0.997} 
& \textbf{0.675} & \textbf{0.718} & \textbf{0.564} & \textbf{0.367} & \textbf{0.571} & \textbf{0.874}
&  \textbf{0.981}	& 0.910 & \textbf{0.962} & 0.993 \\
     & (diff. \%p)  & (5.7) & (2.3) & (0.4) 
& (3.1) & (3.8) & (5.2) & (4.3) & (2.9) & (0.4)
&  (0.6)	& (-0.1) & (0.0) &  (-0.1)
 \\
\bottomrule
\end{tabular}
\end{threeparttable}
}
\caption{Results of previous methods and our model on E-Commerce Corpus, Douban Corpus, and  Ubuntu V1 Corpus. All the results except ours are from the existing literature.  The last row shows the difference between our result and prior state-of-the-art result. }
\label{true_response_res}
\end{table*}

\subsection{Experimental Results}
Table \ref{true_response_res} presents the evaluation results of our model and previous methods on three datasets. As we can see that, {$\mbox{BERT}^+_{TL}$} outperforms the present models on most metrics. Our model outperforms the prior state-of-the-art model by 5.7\% in $\mbox{Recall}_{10}@1$ on E-Commerce Corpus, 3.1\% MAP and 3.8\% MRR on Douban Corpus.  {$\mbox{BERT}^+_{TL}$} also shows a superior ability to the previous fine-tuning methods like {BERT-SL}. Furthermore,  as shown in Table \ref{response_res}, in comparison to the common contrastive learning method {$\mbox{BERT}^+_{DROP}$}, {$\mbox{BERT}^+_{TL}$}  achieves an absolute improvement in $\mbox{Recall}_{10}@1$ by 2.3\% on Douban Corpus and 1\% on E-Commerce Corpus. All these results demonstrate the effectiveness of our model to get better representations and retrieve the best-matched response. Readers can refer to Appendix for training details.




\section{Analysis}
Our model differs in various ways from the previous models: it adds the supervised contrastive objective to the loss function, and it integrates a two-level augmentation technique into the supervised contrastive learning model. We analyze the impact of these components on overall performance.


\begin{table}[htbp]
\centering
\resizebox{0.48\textwidth}{!}{
\begin{tabular}{l|l|cccccc} 
\toprule
                          Data &\diagbox[]{Model}{Metrics} & $\text{\textbf{Recall}}_{10}@1$  & $\text{\textbf{Recall}}_{10}@2$  & $\text{\textbf{Recall}}_{10}@5$  & $\text{\textbf{P}}@1$  & \textbf{MAP}   & \textbf{MRR}  \\ 

\midrule
\multirow{3}{*}{E-Commerce} & $\mbox{BERT}^+$  & 0.903 & 0.973 & 0.998 & - & - & -\\

                            & $\mbox{BERT}^+_{TL}$ w/o CL & 0.914(+1.1\%)      & 0.976(+0.3\%)      & 0.999(+0.1\%)     & -           & -      & -       \\

                            & $\mbox{BERT}^+_{TL}$       & 0.927(+2.4\%)       & 0.974(+0.1\%)      & 0.997(-0.1\%)         & -           & -      & -       \\

\hline
\multirow{3}{*}{Douban}      & $\mbox{BERT}^+$ &  0.342 & 0.558 &  0.870 & 0.535 & 0.657 & 0.700 \\
                            & $\mbox{BERT}^+_{TL}$ w/o CL & 0.351(+0.9\%)      & 0.562(+0.4\%)      & 0.878(+0.8\%)          & 0.546(+1.1\%)       & 0.664(+0.7\%)  & 0.707(+0.7\%)   \\

                            & $\mbox{BERT}^+_{TL}$       & 0.367(+2.5\%)      & 0.571(+1.3\%)      & 0.874(+0.4\%)             & 0.564(+2.9\%)       & 0.675(+1.8\%)  & 0.718(+1.8\%)   \\
\hline
\multirow{3}{*}{Ubuntu}     & $\mbox{BERT}^+$ & 0.907	& 0.959 & 0.992 & - & -\\
                            & $\mbox{BERT}^+_{TL}$ w/o CL & 0.907(+0.0\%)      & 0.960(+0.1\%)      & 0.992(+0.0\%)      & -     & -           & -           \\
                            & $\mbox{BERT}^+_{TL}$        & 0.910(+0.3\%)    & 0.962(+0.3\%)      & 0.993(+0.1\%)  & -     & -           & -        \\

\bottomrule
\end{tabular}}
\caption{The effectiveness of different components on three datasets.}
\label{perf_diff1}
\end{table}

To evaluate the effectiveness of supervised contrastive learning loss, we compare the models with augmentations and contrastive loss ($\mbox{BERT}^+_{TL}$) and models with augmentations but without contrastive loss ($\mbox{BERT}^+_{TL}$ w/o CL).  We use {$\mbox{BERT}^+$}, the BERT-base model after post-training and standard fine-tuning process, as the baseline. Table \ref{perf_diff1} shows that the models with two-level augmentations have already outperformed those without augmentations, and training with contrastive learning loss further boosts the performance. 

\begin{table}[htbp]
\centering
\resizebox{0.46\textwidth}{!}{
\begin{tabular}{l!{\vrule width \lightrulewidth}l!{\vrule width \lightrulewidth}cccccc} 
\toprule
Data                        & \diagbox{Model}{Metrics} &  $\text{\textbf{Recall}}_{10}@1$  & $\text{\textbf{Recall}}_{10}@2$  & $\text{\textbf{Recall}}_{10}@5$  & $\text{\textbf{P}}@1$  & \textbf{MAP}   & \textbf{MRR}   \\ 
\midrule
\multirow{4}{*}{E-Commerce} & $\mbox{BERT}^+$  & 0.903 & 0.973 & 0.998 & - & - & -\\
							& $\mbox{BERT}^+_{DROP}$  & 0.917      & 0.974      & 0.998      & -     & -     & -      \\
                            & $\mbox{BERT}^+_{SR}$    & 0.925      & 0.976      & 0.997      & -     & -     & -      \\
                            & $\mbox{BERT}^+_{STS}$   & 0.919      & \textbf{0.978}      & \textbf{0.999}      & -     & -     & -      \\
                            & $\mbox{BERT}^+_{TL}$    & \textbf{0.927}      & 0.974      & 0.997      & -     & -     & -      \\
\hline
\multirow{4}{*}{Douban}      & $\mbox{BERT}^+$ &  0.342 & 0.558 &  0.870 & 0.535 & 0.657 & 0.700 \\
							& $\mbox{BERT}^+_{DROP}$  & 0.344      & 0.562      & 0.871      & 0.545 & 0.707 & 0.664  \\
                            & $\mbox{BERT}^+_{SR}$    & 0.353      & 0.564      & \textbf{0.878}      & 0.547 & 0.709 & 0.666  \\
                            & $\mbox{BERT}^+_{STS}$   & 0.355      & \textbf{0.575}      & 0.870      & 0.549 & 0.708 & 0.667  \\
                            & $\mbox{BERT}^+_{TL}$    & \textbf{0.367}      & 0.571      & 0.874      & \textbf{0.564} & \textbf{0.718} & \textbf{0.675}  \\ 
\hline
\multirow{4}{*}{Ubuntu}      & $\mbox{BERT}^+$ & 0.907	& 0.959 & 0.992 & - & -\\
							& $\mbox{BERT}^+_{DROP}$  & 0.909      & 0.961      & \textbf{0.993}      & -     & -     & -      \\
                            & $\mbox{BERT}^+_{SR}$    & \textbf{0.910}      & \textbf{0.962}      & \textbf{0.993}      & -     & -     & -      \\
                            & $\mbox{BERT}^+_{STS}$   & \textbf{0.910}      & 0.961      & 0.992      & -     & -     & -      \\
                            & $\mbox{BERT}^+_{TL}$    & \textbf{0.910}      & \textbf{0.962}      & \textbf{0.993}      & -     & -     & -      \\

\bottomrule
\end{tabular}}
\caption{Comparing models with different augmentation techniques on three datasets.}
\label{response_res}
\end{table}

To investigate which augmentation technique in contrastive learning is more beneficial for response selection, we compare different data augmentation approaches, including Dropout (DROP), Sentence Token Shuffling (STS), Sentence Reordering (SR) and Two-Level augmentation (TL) in Table \ref{response_res}. {$\mbox{BERT}^+$} is used as the baseline. {$\mbox{BERT}^+_{DROP}$}, {$\mbox{BERT}^+_{STS}$} and {$\mbox{BERT}^+_{SR}$} are the models fine-tuned using simple supervised contrastive learning loss with DROP, STS and SR as the augmentation strategy. Contrastive learning with standard dropout as noise, is a technique first proposed in SimCSE \cite{gao2021simcse}, and widely used in contrastive learning models in NLP. The results show that the simple method {$\mbox{BERT}^+_{DROP}$} has already achieved consistent improvements across three datasets. For further comparison, we obtain the conclusion that STS and SR are more effective than DROP strategy and in most cases STS is slightly better than SR, indicating the representation space learned with sentence-level augmentation technique is more expressive in a multi-turn dialogue setting. Moreover, we find that the two-level augmentation technique further improves the performance by 0.9\% in MAP and MRR on Douban Corpus. 

\section{Conclusion} 
We propose a new supervised contrastive  learning method and apply it to the task of response selection in multi-turn dialogues. We show that our method outperforms existing approaches on  multiple  datasets:  Ubuntu  V1,  Douban,  and E-Commerce.  Particularly,  we  significantly  boost the performances of the state-of-the-art methods on Douban (5.8\%) and E-Commerce (4.3\%) in terms of $R10@1$.

Our method called two-level supervised contrastive learning makes use of two techniques, Sentence Token Shuffling and Sentence Reordering. We demonstrate the effectiveness of our method through an ablation study. Our method can be potentially applied to other tasks such as reading comprehension and intent classification. We plan to conduct further investigations in the future.

\bibliography{anthology,custom}
\bibliographystyle{acl_natbib}

\appendix
\section{Appendix}
\subsection{Domain Adaptive Post Training}
\label{sec:appendix}


Since most of pre-trained language models (PrLMs) are trained on general corpus (e.g., Wikipedia, Book Corpus \cite{bookcorpus}, Common Crawls \cite{cc}), it is insufficient to have enough supervision of domain-specific words during fine-tuning. To help the model understand domain-specific data, we perform domain adaptive post-training using the original dialogue corpus with pre-training tasks.

Inspired by prior work \cite{han-etal-2021-fine}, we utilize the following procedure to generate training examples to fully exploit the training data.

\begin{itemize}
    \item[1.] Retrieve a random example $(c_i, r_i, y_i)$ where $y_i = 1$, and let dialogue $d_i = c_i \cup \{r_i\}$
    \item[2.] With 50\% possibility, randomly cut a part of $d_i$ as new context $d_i'$ which contains at least 2 utterance, for another 50\% possibility, let $d_i' = d_i$
    \item[3.] Sample the sentence for Next Sentence Prediction (NSP). With 25\% probability, we generate a positive example with the last utterance of $d_i$ as response and the rest as context. For negative examples, 2/3 of the false responses are utterances chosen from corpus at random, and 1/3 are utterances chosen from the same context. We denote this example as $d_i''$
    \item[4.] Convert $d_i''$ to training example $e_i$ in form of BERT input.
    \item[5.] Perform BERT-style whole word masking on $e_i$ for MLM task
\end{itemize}

\subsection{Datasets}
\subsubsection{Ubuntu V1} Ubuntu Corpus V1 is a large multi-turn dialogue dataset extracted from Ubuntu IRC, including technical discussions of the Ubuntu system. We use the version generated by \citet{Xu2016IncorporatingLK}. 
\subsubsection{Douban} Douban Corpus \cite{wu-etal-2017-sequential} is an open-domain Chinese multi-turn dialogue dataset crawled from the popular Douban SNS. The test set of the dataset is human-labeled. 
\subsubsection{E-Commerce} E-commerce Corpus \cite{zhang-etal-2018-modeling} is a Chinese multi-turn dialogue dataset collected from real-world customer services in Taobao, which is the largest e-commerce platform in China.\\\\
The detailed profile of these three data is shown in Table \ref{data_profile}. In the training set of all data, one dialogue context is matched with one correct response and one incorrect response. We can use the false response given by the dataset as a hard negative result.

\begin{table}[tbp]
\begin{tabular}{c|cccc}
\hline
                         & Part  & Ubuntu V1 & Douban & E-Commerce \\ \hline
\multirow{3}{*}{\#pairs} & Train & 1M        & 1M     & 1M         \\ 
                         & Dev   & 500K      & 50K    & 10K        \\  
                         & Test  & 500K      & 6670   & 10K        \\ \hline
\multirow{3}{*}{Pos:Neg} & Train & 1:1       & 1:1    & 1:1        \\ 
                         & Dev   & 1:9       & 1:1    & 1:1        \\  
                         & Test  & 1:9       & 1:4.62     & 1:9        \\ \hline
\end{tabular}
\caption{Data statistics for the Ubuntu, Douban, and E-Commerce Corpus. }
\label{data_profile}
\end{table}

\subsection{Implementation Details}
\subsubsection{Domain Adaptive Training}
We use the English BERT-base model released by Google for the English dataset Ubuntu V1 Corpus and the Chinese BERT model from google for Chinese datasets Douban Corpus and E-Commerce Corpus. For all datasets, we use a batch size of 128, a learning rate of 3e-5 with linear decay, and a warmup ratio of 0.01. We perform domain-adaptive post-training on each dataset for 2.5 million steps. 

\subsubsection{Contrastive Fine-tuning}
For contrastive fine-tuning, we fine-tune each model for 3 epochs and evaluate the result every 2000 steps. For Ubuntu Corpus, we use the original batch size of 48, a learning rate of 2e-5 with linear decay, a hard negative penalty $\alpha=1.0$, and the temperature $\tau=1.0$. For Douban Corpus, we use the original batch size of 43, a learning rate of 1e-5 with linear decay, a hard negative penalty $\alpha=1.0$, and the temperature $\tau=0.05$. For E-Commerce Corpus, we use the original batch size of 43, a learning rate of 2e-5 with linear decay, a hard negative penalty $\alpha=1.0$, and the temperature $\tau=0.05$. For all datasets, we set $\lambda=1.0$ to combine contrastive loss and cross entropy loss.

All codes are implemented based on Huggingfacce Transformer library and pytorch. We perform all experiment on 8 NVIDIA Tesla V100 and fp16 mix precision calculation is applied to all experiments.

\subsection{Related Work}
\subsubsection{Pre-trained Language Models}
Pretraining and fine-tuning have become a new paradigm of natural language processing and understanding. GPT \cite{Radford2019LanguageMA}, BERT \cite{devlin-etal-2019-bert}, RoBERTa \cite{liu2020roberta}, XLNet \cite{NEURIPS2019_dc6a7e65}, and ELECTRA \cite{Clark2020ELECTRA:} are prevailing pre-trained language models. GPT was trained using uni-directional language model objective, while BERT was trained in a bidirectional way using NSP and MLM tasks. RoBERTa trained the language model with more data and removed the NSP objective from the training task. ELECTRA introduced novel generator-discriminator architecture for language models to improve training efficiency. StructBERT \cite{Wang2020StructBERT:} incorporated the word structural objective and sentence structural objective to leverage language structures at the word and sentence levels.

\subsubsection{Multi-Turn Dialogue Response Selection}
Before the advent of pre-trained language models, researchers used CNNs and various RNNs to process the dialogue. Early work focused on the word or sentence matching. \citet{lowe-etal-2015-ubuntu} first employed LSTM and CNNs in response selection task and proposed Ubuntu V1 dataset. Later work usually encoded and matched dialogues at different levels to gain better performance. \citet{wu-etal-2017-sequential} proposed sequential matching model and Douban Dataset, \citet{zhang-etal-2018-modeling} proposed the deep utterance aggregation and E-Commerce dataset. After self-attention and Transformer architecture were introduced, studies began to leverage self-attention architecture such as Deep Attention Network \cite{zhou-etal-2018-multi} and Multi Representation Fusion Network to perform matching. Interaction over Interaction \citet{tao-etal-2019-one} designed deep interaction module and Multi-hop Selection Network \citet{yuan-etal-2019-multi} focused on selecting valid context for matching.

After PrLMs are proposed, they are directly applied to response selection since dialogues can also be considered as texts. To further improve the performance of PrLMs on dialogue response selection, researchers attempt to add components to PrLMs. SA-BERT \cite{sa-bert} added speaker embedding to the model, and MDFN \cite{Liu2021FillingTG} added more dialogue-aware response interaction module. Domain Adaptive Post Training is also an effective approach for dialogue response selection, and this approach was first proposed by \citet{bert-vft}. Previous state-of-the-art model BERT-FP \cite{han-etal-2021-fine} used a fine-grained post-training approach to improve the performance and Structural Pre-training for Dialogue Comprehension was used. Multi-Task Learning is also an effective way, UMS \cite{Whang_Lee_Oh_Lee_Han_Lee_Lee_2021} and BERT-SL \cite{Xu2021LearningAE} are two examples of applying self-supervised multi-task learning to dialogue response selection.

\subsubsection{Contrastive Learning}
Contrastive learning is first introduced by \citet{1640964} and re-emerges in Computer Vision (CV) and Natural Language Processing (NLP) recently. In computer vision, SimCLR \cite{chen2020simple} proposed a simple framework that can be applied to CV and \citet{khosla2020supervised} proved the effectiveness of contrastive learning under supervised scenarios. In NLP, CERT \cite{Fang2020CERTCS} was proposed to improve language understanding, while ConSERT \cite{yan2021consert}, DeCuLTR \cite{giorgi-etal-2021-declutr}, and SimCSE \cite{gao2021simcse} were proposed to give good representation of sentence. In other tasks, SimCLS  \cite{liu-liu-2021-simcls} applied contrastive learning in summarization, \citet{pan2021improved} performed contrastive learning in text classification with adversarial training and \citet{gunel2021supervised} proved that fine-tuning PrLMs can also improve the performance.

\end{document}